\documentclass[letterpaper]{article} 
\usepackage{aaai2026}  
\usepackage{times}  
\usepackage{helvet}  
\usepackage{courier}  
\usepackage[hyphens]{url}  
\usepackage{graphicx} 
\def\UrlFont{\rm}  
\usepackage{natbib}  
\usepackage{caption} 
\usepackage[ruled,vlined]{algorithm2e}
\usepackage{caption}
\usepackage{subcaption}
\usepackage{amsmath}
\usepackage{amssymb}
\DeclareMathOperator*{\argmax}{arg\,max}

\SetKwInput{KwInput}{Input}                
\SetKwInput{KwOutput}{Output}              
\SetKwInput{KwInit}{Initialization}
\usepackage{makecell}
\usepackage{multirow}
\usepackage[table]{xcolor}
\usepackage{bbding}
\usepackage{booktabs}

\usepackage{algorithm}
\usepackage{algorithmic}

\usepackage{newfloat}
\usepackage{listings}
\DeclareCaptionStyle{ruled}{labelfont=normalfont,labelsep=colon,strut=off} 
\floatstyle{ruled}
\newfloat{listing}{tb}{lst}{}
\floatname{listing}{Listing}
\title{Concepts from Representations: Post-hoc Concept Bottleneck Models via \\Sparse Decomposition of Visual Representations}

\author {
    Shizhan Gong\textsuperscript{\rm 1},
    Xiaofan Zhang\textsuperscript{\rm 2},
    Qi Dou\textsuperscript{\rm 1}
}
\affiliations {
    \textsuperscript{\rm 1}The Chinese University of Hong Kong, Hong Kong, China\\
    \textsuperscript{\rm 2}Shanghai Jiao Tong University, Shanghai, China\\
    szgong22@cse.cuhk.edu.hk, xiaofan.zhang@sjtu.edu.cn, qidou@cuhk.edu.hk
}

\usepackage{bibentry}

\begin{document}

\maketitle
\begin{abstract}
Deep learning has achieved remarkable success in image recognition, yet their inherent opacity poses challenges for deployment in critical domains. Concept-based interpretations aim to address this by explaining model reasoning through human-understandable concepts. However, existing post-hoc methods and ante-hoc concept bottleneck models (CBMs), suffer from limitations such as unreliable concept relevance, non-visual or labor-intensive concept definitions, and model/data-agnostic assumptions. This paper introduces  \textbf{P}ost-hoc \textbf{C}oncept \textbf{B}ottleneck \textbf{M}odel via \textbf{Re}presentation \textbf{D}ecomposition (\textbf{PCBM-ReD}), a novel pipeline that retrofits interpretability onto pretrained opaque models. PCBM-ReD automatically extracts visual concepts from a pre-trained encoder, employs multimodal large language models (MLLMs) to label and filter concepts based on visual identifiability and task relevance, and selects an independent subset via reconstruction-guided optimization. Leveraging CLIP’s visual-text alignment, it decomposes image representations into linear combination of concept embeddings to fit into the CBMs abstraction.  Extensive experiments across 11 image classification tasks show PCBM-ReD achieves state-of-the-art accuracy, narrows the performance gap with end-to-end models, and exhibits better interpretability. 
\end{abstract}    
\begin{links}
    \link{Code}{https://github.com/peterant330/PCBM_ReD}
\end{links}
\section{Introduction}
\label{sec:intro}
Deep learning has made significant strides in various image recognition tasks. However, the complexity of these models, which is often necessary to achieve high accuracy, leads to an opaque behavior. This limits the broader application in critical fields such as medical imaging analysis~\cite{gong2025concepts} and autonomous driving~\cite{omeiza2021explanations}. Concept-based interpretation is a subfield of explainable artificial intelligence that aims to use human understandable concepts to explain the model behaviors. Concept-based methods usually represent the semantic features learned by the network with concept labels, so that human can understand which features are responsible for the final predictions. 

\begin{figure}[t]
\centering
\includegraphics[width=0.48\textwidth]{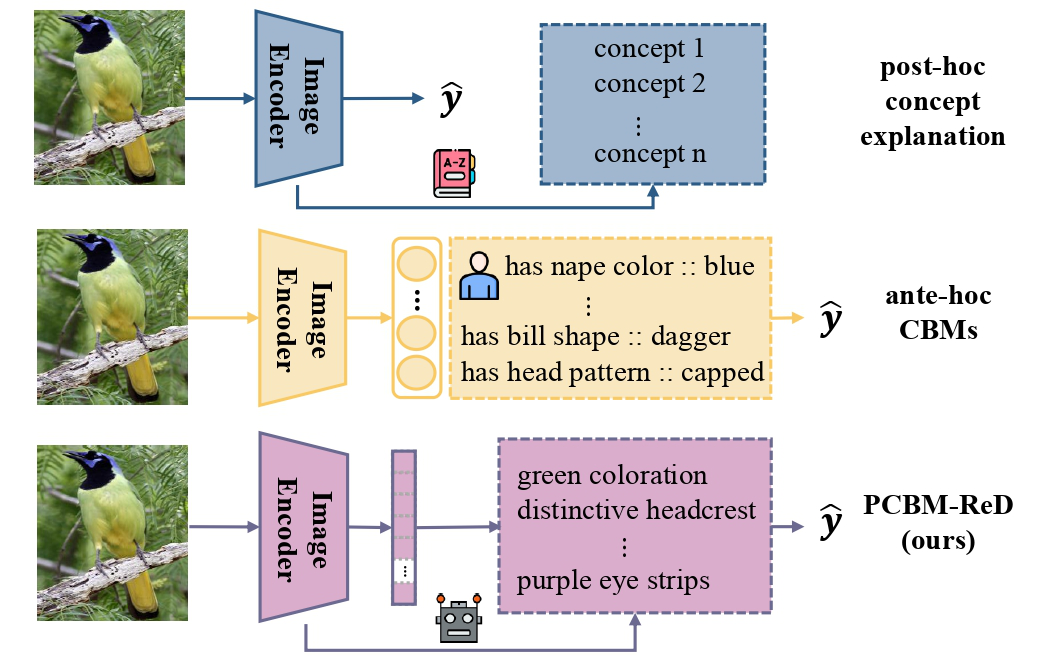}
\caption{PCBM-ReD extracts concepts from the pre-trained image encoder and reconstruct the visual representation with the concepts, which gives faithful interpretation and can take the best advantage of the encoder's representation power.}
\label{teaser}
\end{figure}

Concept-based methods can be divided into post-hoc methods and ante-hoc approaches (Fig.~\ref{teaser}). Post-hoc methods associate latent space
representations to human-understandable concepts, either supervised with user-defined concepts~\cite{bau2017network,fong2018net2vec} or by unsupervised pattern recognition~\cite{ge2021peek,vielhaben2023multi}. However, these methods face some major limitations. Firstly, there is no guarantee that the extracted concepts faithfully reflect the reasoning process of the network. Additionally, there may be no intuitive and human-understandable causal relationship between the extracted concepts and targets, making it difficult for users to understand the mechanism of the network. Moreover, automated concepts mining process can generate a great number of concepts with low semantic value, which are difficult to be labeled by humans~\cite{kim2018interpretability}. 

Concept Bottleneck Models (CBMs) \cite{koh2020concept} are ante-hoc approaches that integrate concept associations into neural networks through modifications in model design or training. They predict a set of interpretable concepts and use a linear function to generate final predictions based on these concepts. This structure allows users to trace the links between concepts and class labels, facilitating error correction. CBMs rely on user-defined concepts, which can be either manually crafted by experts \cite{koh2020concept,yun2022vision} or generated by large language models (LLMs) \cite{yang2023language,oikarinen2023label}. However, the current CBMs exhibit several weaknesses: human-crafted concepts can be labor-intensive and may lack comprehensive coverage \cite{koh2020concept}, while LLMs-generated concepts often include non-visual features (e.g., the taste of food or bird behavior, which are not visually inferred) \cite{yang2023language}. Additionally, these pre-defined concepts are data-independent and model-agnostic, which can hinder their effectiveness, especially if a dataset is biased towards certain subpopulation or if the image encoder cannot capture certain features (e.g., color or spatial relationships \cite{kamath2023whats}). Furthermore, current methods do not guarantee the independence of generated concepts, which is essential for effective interventions in CBMs. These limitations hinder CBMs from achieving complete feature space interpretability \cite{kim2018interpretability}.

In this work, we introduce a novel pipeline that retrofits interpretability onto pretrained opaque foundation models, called \textbf{P}ost-hoc \textbf{C}oncept \textbf{B}ottleneck \textbf{M}odel via \textbf{Re}presentation \textbf{D}ecomposition (\textbf{PCBM-ReD}). Given a pre-trained image encoder, we first apply automatic concept extraction~\cite{fel2023holistic} to mine the generalizable features encoded within the foundation models. To further ensure they are visually-identifiable, human-understandable, and task-related, we use multimodal large language models (MLLMs) to label concepts by summarizing descriptions of top-activated images for each concept and scoring them based on prior knowledge of the task. We then propose a reconstruction-guided concept selection algorithm that selects a subset of concepts, whose embedding spans are independent and completely define the visual representation space. For image-concept association, we utilize the visual-text alignment property of CLIP~\citep{radford2021learning}, which allows us to decompose the visual representation into a weighted sum of the concepts' text embeddings. By eliminating residual terms and fitting a linear function to the reconstructed representation, we can develop a model that adheres to the abstractions of CBMs while preserving the representational power of the original opaque visual encoder. 

We conduct comprehensive experiments to demonstrate the efficacy of our proposed method. We evaluate the model performance on 11 image classification tasks, including common object recognition, fine-grained types, texture, and action classification, as well as domain-specific tasks such as medical diagnosis and satellite image object recognition. Our main finding is that the model achieves state-of-the-art classification accuracy, with a reduced gap compared to end-to-end models,  compared to existing CBMs. Furthermore, as we aim to mimic the behavior of the end-to-end model using CBM, our approach exhibits similar properties, including zero-shot and few-shot capabilities. Additionally, we conduct human evaluation to demonstrate the improved interpretability of PCBM-ReD. Our main contributions include:
\begin{itemize}
\item  We propose a data-driven scheme for creating and selecting concepts that align with the data distribution and the image encoder's representation capabilities.
\item  We propose constructing CBMs by leveraging CLIP's visual-text alignment to sparsely decompose the visual representation into concept embeddings.
\item We demonstrate that PCBM-ReD achieves SOTA classification accuracy across various tasks, along with better interpretability and robust zero/few-shot capabilities.
\end{itemize}

\section{Related Work}

\noindent \textbf{Explanations in Computer Vision.}
Explainable computer vision primarily falls into two categories: post-hoc explanations and interpretable models by design. Post-hoc explanation methods generate saliency maps to identify which input features influence the decisions of neural networks~\cite{simonyan2013deep,selvaraju2017grad,gong2024structured,gong2025boosting}. However, these methods do not guarantee that the explanations faithfully reflect the model's reasoning process~\cite{rudin2019stop}. In contrast, interpretable models by design ensure that explanations align with the models' reasoning~\cite{chen2019looks,nauta2021neural,ma2024looks,tan2024post}. Our work builds upon CBMs~\cite{koh2020concept}, a type of interpretable models.

\noindent \textbf{Concept Bottleneck Models.}
CBMs~\cite{koh2020concept,zarlenga2022concept,kim2023probabilistic,gong2025concepts} predict outcomes by linearly combining an intermediate layer of human-understandable attributes. The original CBM relies on handcrafted and manually annotated attributes. CompDL~\cite{yun2022vision} replaces manual annotations with CLIP scores, but still depends on concepts designed by human. LaBo~\cite{yang2023language} and label-free CBM~\cite{oikarinen2023label} further automate concept generation using LLMs. Early CBMs often underperformed compared to end-to-end models. To improve it, Post-hoc CBM~\cite{yuksekgonul2022post} introduces a residual connection from image features to predictions, while Res-CBM~\cite{shang2024incremental} approximates this connection by incrementally adding new concepts. OpenCBM~\cite{tan2024explain} detects missing concepts from an open vocabulary. Our method reduces residuals from the beginning by mining concepts that can well reconstruct the image representations.

\begin{figure*}[t]
\centering
\includegraphics[width=0.95\textwidth]{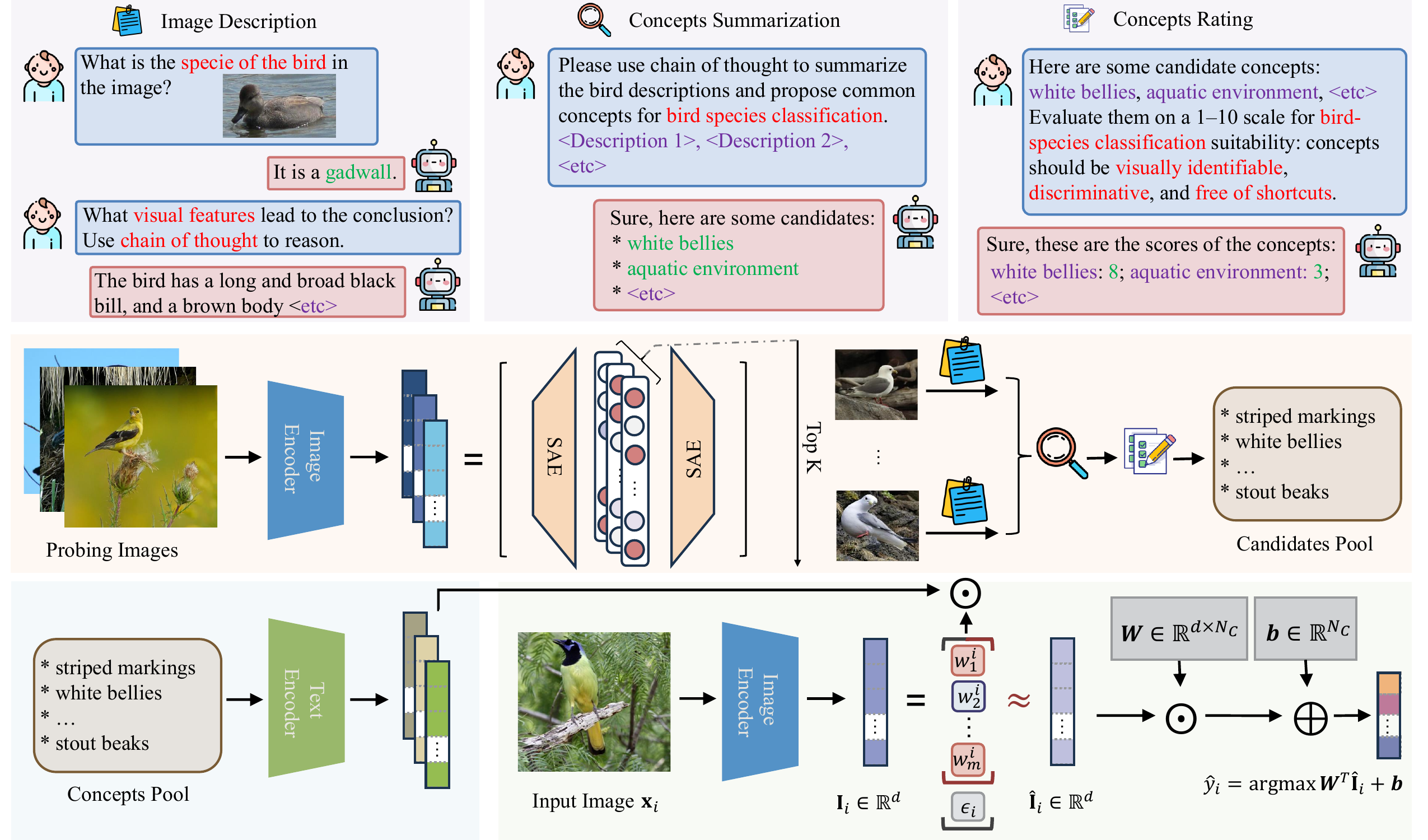}
\caption{We present an overview of \textbf{P}ost-hoc \textbf{C}oncept \textbf{B}ottleneck \textbf{M}odel via \textbf{Re}presentation \textbf{D}ecomposition (\textbf{PCBM-ReD}). First, we extract concepts from the learned representation, and use MLLMs to summarize and score the concepts. Second, we apply a concept selection algorithm to choose concepts and construct the bottleneck. Third, we perform sparse decomposition and reconstruct the image embedding by concepts. A linear layer is trained to predict the targets with the fitted embedding.}
\label{method}
\end{figure*}

\noindent \textbf{CLIP Interpretation.}
CLIP~\citep{radford2021learning} utilizes contrastive learning on a large dataset of paired text and images to link images with their textual descriptions. This approach effectively groups similar concepts together and shows strong performance in downstream tasks, such as zero-shot classifications~\cite{saha2024improved}.
Several studies focusing on the interpretation of CLIP have shown that its homogeneous feature space allows the visual embeddings to be decomposed into multiple concepts represented by concepts' text embeddings~\citep{moayeri2023text,chen2023stair,gandelsman2023interpreting,gandelsman2024neurons}. This provides the theoretical foundation for our method, which constructs CBMs by directly decomposing visual representations into concepts representation, thus better preserving predictive power.

\section{Method}

\begin{table*}[tp]
\centering
\renewcommand\arraystretch{1}
    
\begin{center}
\setlength{\tabcolsep}{1mm}
\fontsize{9}{10}\selectfont
\begin{tabular}{cccccccccccccc}
\toprule
Method &\rotatebox[origin=c]{45}{Interpret.}& \rotatebox[origin=c]{45}{ImageNet} & \rotatebox[origin=c]{45}{CIFAR10}&\rotatebox[origin=c]{45}{CIFAR100}&\rotatebox[origin=c]{45}{FOOD}&\rotatebox[origin=c]{45}{Aircraft}&\rotatebox[origin=c]{45}{Flower}&\rotatebox[origin=c]{45}{CUB}&\rotatebox[origin=c]{45}{UCF}&\rotatebox[origin=c]{45}{DTD}&\rotatebox[origin=c]{45}{HAM}&\rotatebox[origin=c]{45}{RESISC}&Average\\
\midrule
\multicolumn{14}{c}{\textbf{Fully-supervised Setting}}\\
\midrule
Linear Probe &\XSolidBrush & 83.90&98.10&87.48&93.17&64.03&99.45&84.54&90.67&81.68&83.18&94.98&87.38\\
\midrule
LaBo &\Checkmark &83.97 &97.75 &86.04 &92.45 &61.42 &99.35 &81.90 &90.11  &77.30  &\textbf{81.39} &91.22  & 85.72 
\\
Res-CBM &\Checkmark & 82.98 &97.77 &83.01& 90.17& 54.67& 97.85& 79.27& 88.37& 75.77 &75.72 & 91.67 &83.39
\\
V2C-CBM &\Checkmark & 84.15 &98.03 &86.41& 92.84& 60.71& 98.94& 83.12& -& 78.49 &81.12 & 92.86 &-\\
\midrule 
PCBM-ReD (ours)&\Checkmark&\textbf{84.48} &\textbf{98.05}&\textbf{87.27} &\textbf{93.16} &\textbf{62.95} &\textbf{99.39} &\textbf{84.80} &\textbf{90.38} &\textbf{81.44} &\textbf{81.39} &\textbf{93.31} & \textbf{86.97} \\
\midrule
\multicolumn{14}{c}{\textbf{Zero-shot Classification Setting}}\\
\midrule
CLIP & \XSolidBrush & 72.90&95.57&78.28&90.91&31.77&79.46&62.19&75.31&57.09&58.21&64.87&69.69\\
PCBM-ReD (ours)&\Checkmark &72.89&95.56&78.24&90.91&32.01&79.58&62.03&75.36&57.09&58.51&64.87&69.73\\
\midrule
CuPL& \XSolidBrush &73.45&95.82&78.59&91.23&35.43&80.80&64.43&76.00&62.65&57.91&71.88&71.65\\
PCBM-ReD + CuPL (ours)& \Checkmark&73.43&95.80&78.59&91.22&35.52&80.67&64.62&75.91&62.77&58.11&71.88&71.68 \\
\bottomrule
\end{tabular}
\end{center}
\caption{Test accuracy (\%) of PCBM-ReD on 11 image classification benchmarks. We report performance of both fully-supervised setting and zero-shot setting. For fully-supervised setting, we compare PCBM-ReD with linear probe, LaBo and Res-CBM. For zero-shot setting, we utilize two strategies, i.e., vanilla CLIP and CuPL. CLIP ViT-L/14 is used as the backbone.}
\label{tab:main1}
\end{table*}

Consider a training set of image-label pairs $\mathcal{D} = \{(\mathbf{x},y)\in \mathbb{R}^{d_x}\times\mathcal{Y}\}$, and a bottleneck $C$ made of $N_C$ concepts, $C:=\{c_1, c_2,\cdots, c_{N_C}\}$. CBM make prediction by composing two functions, $\hat{y} = f(g(\mathbf{x}))$, where $g:\mathbb{R}^{d_x}\rightarrow \mathbb{R}^{N_C}$ maps $\mathbf{x}$ to a score for each element of the bottleneck, and $f:\mathbb{R}^{N_C} \rightarrow \mathcal{Y}$ makes the final prediction on the label space given the concept scores. Fig.~\ref{method} presents an overview. 

\subsection{Data-driven Concept Discovery}
\label{sec:3.1}
Most existing methods for constructing bottlenecks rely on either handcrafted concepts~\cite{koh2020concept} or concepts generated by LLMs~\cite{yang2023language}. However, these concept creation processes are typically independent of the training data or the image encoder, which can result in suboptimal concepts that fail to capture variations in specific data distributions or are challenging for the visual encoder to distinguish. To overcome these limitations, we propose an approach that automatically extracts concepts from a pre-trained image encoder. We leverage the pre-trained multi-modal alignment model, CLIP~\cite{radford2021learning}, which consists of an image encoder $\mathcal{I}$ and a text encoder $\mathcal{T}$. The image encoder transforms an image $\mathbf{x}_i$ into a representation $\mathbf{I}_i = \mathcal{I}(\mathbf{x}_i) \in \mathbb{R}^d$. If the encoder effectively captures structured patterns, its latent space should be disentangled into subspaces representing distinct concepts~\cite{bau2017network, oikarinen2022clip}. To do so, we apply sparse autoencoder (SAE)~\cite{bricken2023monosemanticity}, which enables us to represent the visual embedding $\mathbf{I}_i$ with the concept bank atoms $\mathbf{u}_i \in \mathbb{R}^k$, such that $\mathbf{I}_i \approx \mathbf{V} \mathbf{u}_i$. The SAE represents $\mathbf{u}_i$ by a neural network $\psi$ (i.e., $\mathbf{u}_i = \psi(\mathbf{I}_i)$) and enforces sparsity on $\mathbf{u}_i$~\cite{fel2023holistic}, known to effectively address the challenge of polysemanticity~\cite{bricken2023monosemanticity}. Other dictionary learning paradigms are also applicable.
 
 Ideally, each column of $\mathbf{V}$ represents a concept and the corresponding value of $\mathbf{u}_i$ reflects the significance of the concept in $\mathbf{I}_i$. To transform the concepts into human-understandable form, we propose a MLLM-based pipeline for labeling and scoring the concepts. We sample a reasonable number of images from the training data as probing images, and obtain their corresponding $\mathbf{u}_i$. Then for each concept, we select the top $K$ images with largest concept scores. We then prompt the MLLMs with chain-of-thought to summarize the concept. As shown in Fig.~\ref{method}, for each image, we prompt MLLMs to describe the visual features that can help identify the category of the image. After collecting image descriptions for the top $K$ images, we ask LLMs to summarize these descriptions and generate candidate concepts that is useful for a specific classification problem. This process results in a large number of candidate concepts, which may or may not be good concepts for the classification tasks. We then ask LLMs to score these concepts. We prompt the LLMs to only assign high scores to concepts that are visually identifiable, discriminatory, and free of shortcuts (e.g., concepts describing the background). We filtered out candidates with low scores, leaving only high-quality concepts.
\subsection{Reconstruction-guided Concept Selection}
\label{sec:3.2}
The LLM-based scoring ensures that the resulting concepts are both human-understandable and relevant to specific tasks. However, the use of a limited sample size for summarizing these concepts can result in noisy outputs with low coverage of the overall dataset. Moreover, the candidate pool may include overlapping or repetitive concepts. To address these issues, it is crucial to select a subset of important and independent concepts that can still comprehensively cover the entire representation space.

We take a set of $N$ probing images sampled from the training data, with $\mathbf{I}_1, \cdots, \mathbf{I}_N$ denoting their image embeddings in the joint text-image space. Given a collection of concepts $\mathcal{C}$, we can reconstruct the image representations by linear combination of the text representations of the concepts within the concept set (denoted as $\mathbf{R}(\mathcal{C})\in\mathbb{R}^{M_\mathcal{C}\times d}$). We further define the reconstruction error to be the Frobenius norm of the difference between the original and the reconstructed image embedding. Our goal is to identify the optimal subset that minimize the reconstruction error:
\begin{equation}
\min_\mathcal{C} \sum_{i=1}^N \min_{\mathbf{\beta}_i(\mathcal{C})} \Vert \mathbf{I}_i - \mathbf{R}(\mathcal{C})^T\mathbf{\beta}_i(\mathcal{C}) \Vert_F^2,
\end{equation}
where $\mathbf{\beta}_i(\mathcal{C})$ is the coefficients of linear combination for reconstructing $\mathbf{I}_i$.
While the subset selection is a discrete optimization problem without close-form solution, we propose a greedy algorithm that select concepts step-wisely. Additionally, although the inner optimization of coefficients $\mathbf{\beta}_i$ has analytical solution, we need to solve the optimization problem for each concept within the set $\mathcal{C}$, which can be computational-intensive when the size of $\mathcal{C}$ is large. We propose an algorithm for efficient computation, as illustrated in Alg.~\ref{alg:alg}. Detailed explanation of the algorithm can be found in Appendix. The algorithm can incrementally select new concepts, minimizing reconstruction error to the greatest extent, while ensuring that the newly added concepts are linearly independent from the existing ones. The algorithm stops when the selected concepts reach a pre-defined value, or when all new concepts are linearly dependent on the existing concepts. It is important to note that, unlike the selection algorithms presented in previous work~\cite{chen2019looks,yang2023language}, this selection scheme is entirely unsupervised, making it suitable for zero/few-shot applications.

\begin{algorithm}[t]
\small
\DontPrintSemicolon
  \KwInput{Image embedding for $N$ images stacked as rows in a matrix $\mathbf{X}\in \mathbb{R}^{N\times d}$, a pool of $M$ concepts $\{c_i\}_{i=1}^M$, their corresponding text representations $\{\mathcal{T}(c_i)\}_{i=1}^M$,   selected concepts size $m$,  identity matrix $\mathbf{E}$}
  
  \KwOutput{A set of selected candidates $\mathcal{C}$ }

  \KwInit{$\mathcal{C} \leftarrow \phi$, $\mathcal{C}_0 \leftarrow \{c_i\}_{i=1}^M$}
  $c^* \leftarrow \arg \max_{c\in\mathcal{C}_0} \left\Vert\mathbf{X}\cdot\left(\mathbf{E} -\frac{\mathcal{T}(c)\mathcal{T}(c)^T}{\mathcal{T}(c)^T\mathcal{T}(c)}\right)\right \Vert_F^2$

$\mathcal{C} \leftarrow \mathcal{C} \cup \{c^*\}$, $\mathcal{C}_0 \leftarrow \mathcal{C}_0 \setminus \{c^*\}$
  
  \For{i in $[2,...,m]$}
    {   
    $\mathbf{P} \leftarrow \mathbf{R}(\mathcal{C})(\mathbf{R}(\mathcal{C})^T\mathbf{R}(\mathcal{C}))^{-1}\mathbf{R}(\mathcal{C})^T$

    \For{c in $\mathcal{C}_0$}
    { $z \leftarrow \mathcal{T}(c)^T(\mathbf{E}-\mathbf{P})\mathcal{T}(c)$
    
    \eIf{$z = 0$}
    {
        $\mathcal{C}_0 \leftarrow \mathcal{C}_0 - \{c\}$
    }{
    $\mathbf{Q} \leftarrow \mathcal{T}(c)\mathcal{T}(c)^T/z$

    $\mathbf{L}(c) \leftarrow \mathbf{P}\mathbf{Q}\mathbf{P}-\mathbf{Q}\mathbf{P}-\mathbf{P}\mathbf{Q}+\mathbf{P}+\mathbf{Q}$
    }}
\If{$\mathcal{C}_0 = \phi$ }
    {
       \textbf{break}
    }
    $c^{*} \leftarrow \arg \min_{c\in \mathcal{C}_0}\Vert\mathbf{X}\left(\mathbf{E} -\mathbf{L}(c)\right)\Vert_F^2$
    
    $\mathcal{C} \leftarrow \mathcal{C} \cup \{c^*\}$, $\mathcal{C}_0 \leftarrow \mathcal{C}_0 \setminus \{c^*\}$
    }
\caption{Concepts Selection Algorithm}
\label{alg:alg}
\end{algorithm}

\subsection{Post-hoc Class-concept Association}
\label{sec:3.3}
\noindent \textbf{Concept Scores Assignment.}
Multimodal learning enables the alignment of representations from different modalities into a joint space. Several studies~\cite{gandelsman2023interpreting,gandelsman2024neurons} have demonstrated that the image embeddings of CLIP can be represented as a weighted sum of text representations. Therefore, rather than relying on manual annotations or text-image similarity scores to generate concept score supervision, we propose to directly decompose the image representation into concept-related directions within the joint representation space. To ensure high interpretability, we aim for this decomposition to be sparse, utilizing only a few key concepts to explain the image representation. Mathematically, we express this as:
\begin{equation}
\mathbf{I}_i = \hat{\mathbf{I}}_i+\epsilon_i = \sum_{j=1}^m w_j^i \mathbf{c}_j + \epsilon_i,
\end{equation}
where $\mathbf{c}_j$ is the embedding of concept $c_j$ and $\epsilon_i$ is residue.  We apply a sparse coding algorithm (e.g., orthogonal matching pursuit~\cite{pati1993orthogonal}) to approximate $\mathbf{I}_i $ as the sum above, where only $n$ of the $w^i_j$ are non-zero, for some $n < m$.

\noindent \textbf{Label Predictor.} After decompose each image embedding $\mathbf{I}_i$ into the sum of concept embedding, we discard the residue and retain only the fitted representation $\hat{\mathbf{I}}_i$. We then fit a linear layer to predict the class label from the fitted representation, expressed as $\hat{y}_i = \argmax{(\mathbf{W}^T\hat{\mathbf{I}}_i + \mathbf{b}})$.  This function can be reformulated as $\hat{y}_i = \argmax{(\sum_{j=1}^m w^i_j \mathbf{W}^T\mathbf{c}_j}+\mathbf{b})$. As a result, it satisfies the CBM abstraction, with the coefficients $[w_1^i, \cdots, w_m^i]$ representing the concept scores, and $\mathbf{W}^T[\mathbf{c}_1, \cdots, \mathbf{c}_m]$ serving as the class-concept weight matrix. Unlike traditional CBMs that train a sparse class-concept weight matrix to improve the model interpretability, our method enforces sparsity on the concept score side by applying sparse decomposition to the visual representation.

\noindent \textbf{Weight Matrix Initialization.}  Since $\hat{\mathbf{I}}_i$ is an approximation of $\mathbf{I}_i$, it shares similar properties with $\mathbf{I}_i$. One advantage of $\mathbf{I}_i$ is its zero-shot capability, stemming from the alignment of image and text representations. Ideally, $\hat{\mathbf{I}}_i$ would exhibit similar zero-shot ability. To better leverage this zero-shot prior, we propose to initialize $\mathbf{W}$ with the text embeddings of the prompt ``\texttt{This is a photo of [cls]}''.

\begin{table}[tp]
\centering
\renewcommand\arraystretch{1}
    
\begin{center}
\setlength{\tabcolsep}{1mm}
\fontsize{9}{10}\selectfont
\begin{tabular}{cccccc}
\toprule
Method & \rotatebox[origin=c]{50}{Interpret.}& \rotatebox[origin=c]{50}{CIFAR10}&\rotatebox[origin=c]{50}{CIFAR100}&\rotatebox[origin=c]{50}{CUB}&Average\\
\midrule
Linear Probe & \XSolidBrush &88.80  &70.10 &72.14 &	77.01  \\
\midrule
Original CBM&\Checkmark&- &-&65.13 &-\\
CompDL&\Checkmark&- &-&54.19 &- \\
PCBM&\Checkmark&84.50&56.00&63.63&68.04\\
Label-free CBM&\Checkmark& 86.40 &65.13 &62.40 &71.31\\
CDM&\Checkmark&86.50&67.60&\textbf{72.26}&75.45\\
DN-CBM&\Checkmark& 87.60 &67.50 &68.38 &74.49\\
VLG-CBM&\Checkmark& \textbf{88.63} &66.48 &66.03 &73.71 \\
\midrule
PCBM-ReD (ours)&\Checkmark&88.61 &\textbf{70.03} &72.01 &\textbf{76.88} \\
\bottomrule
\end{tabular}
\end{center}
\caption{Test accuracy (\%) comparison between PCBM-ReD and baselines for fully-supervised setting. We use CLIP RN50 as the backbone for all methods.}
\label{tab:main2}
\end{table}

\begin{figure}[t]
\centering
\includegraphics[width=0.48\textwidth]{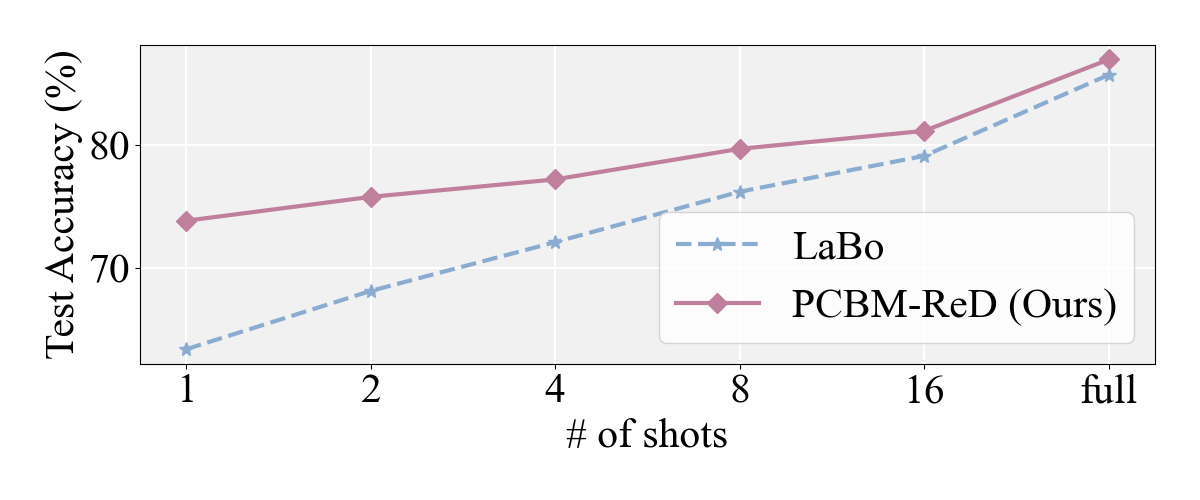}
\caption{Few-shot test accuracy (\%) comparison. Average test accuracy on 11 datasets is reported. Shot means the number of labeled images for each class.}
\label{few_shot}
\end{figure}

\begin{figure*}[t]
\centering
\includegraphics[width=\textwidth]{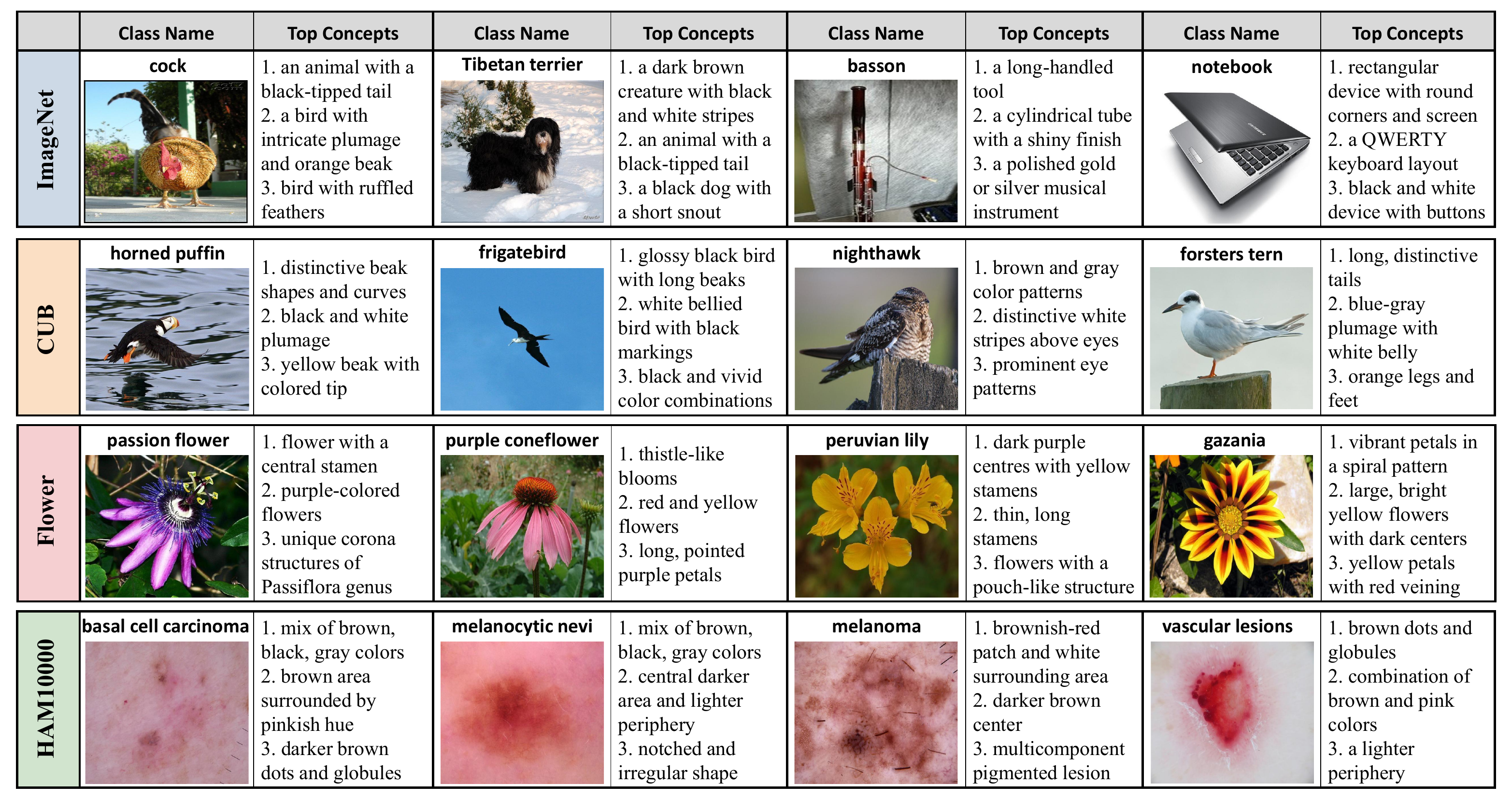}
\caption{Several example explanations generated by PCBM-ReD. The examples are sampled from the test set of 11 datasets, which have correct predictions. We also show their corresponding top concepts that contribute the most to the logits.}
\label{explanation}
\end{figure*}

\begin{figure}[t]
\centering
\includegraphics[width=0.48\textwidth]{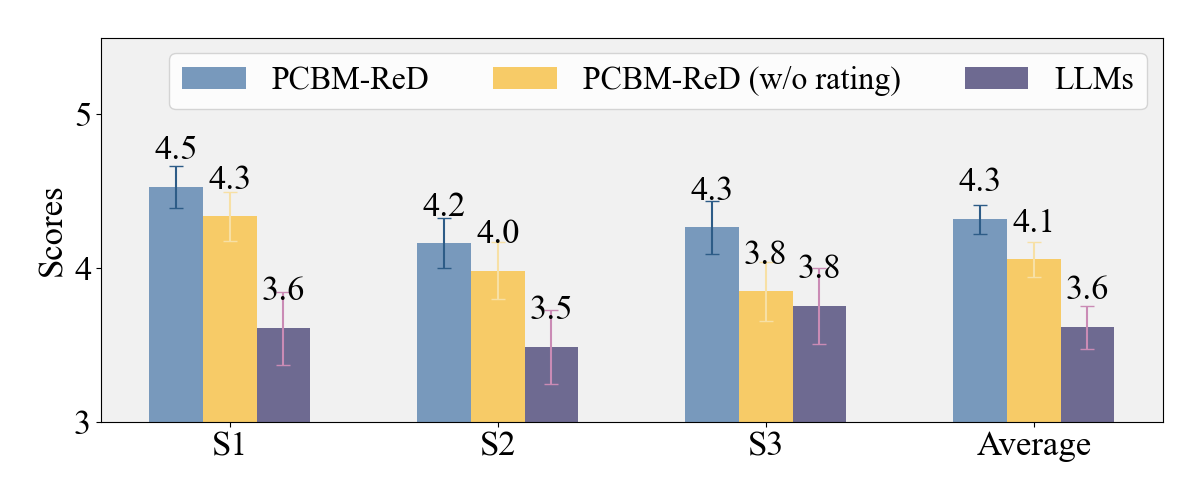}
\caption{Human evaluation. Volunteers rate the explanation on a scale of 1 to 5 (5 = very agree). \textbf{S1}: The explanations are visually identifiable features. \textbf{S2}: The explanations faithfully describe the image. \textbf{S3}: There is a causal relationship between the explanation and the prediction. }
\label{user_study}
\end{figure}
\section{Experiments}
\subsection{Dataset and Baselines}
Following the benchmark proposed by~\cite{yang2023language}, we use 11 image classification datasets spanning a diverse set of domains, including (1) Common objects: ImageNet~\cite{deng2009imagenet}, CIFAR-10 and CIFAR-100~\cite{krizhevsky2009learning}; (2) Fine-grained objects: Food-101~\cite{bossard2014food}, FGVC-Aircraft~\cite{maji2013fine}, Flower-102~\cite{nilsback2008automated}, CUB-200-2011~\cite{wah2011caltech}; (3) Actions: UCF-101~\cite{soomro2012ucf101}; (4) Textures: DTD~\cite{cimpoi2014describing}; (5) Skin tumors: HAM10000~\cite{tschandl2018ham10000}, and (6) Satellite images: RESISC45~\cite{cheng2017remote}. We use the same train/dev/test splits as~\cite{yang2023language} for all datasets. For all experiments, we train on the training set and report the test accuracy. We compare our model, PCBM-ReD, with end-to-end linear probing, as implemented from CLIP~\cite{radford2021learning}, as well as several CBMs, including 
original CBM~\cite{koh2020concept}, PCBM~\cite{yuksekgonul2022post}, CompDL~\cite{yun2022vision}, label-free CBM~\cite{oikarinen2023label}, LaBo~\cite{yang2023language}, Res-CBM~\cite{shang2024incremental}, CDM~\cite{panousis2023sparse}, DN-CBM~\cite{rao2024discover}, V2C-CBM~\cite{he2025v2c}, and VLG-CBM~\cite{srivastava2024vlg}.

\subsection{Implementation Details}
We use Llama-3.2-11B-Vision-Instruct to generate image descriptions and use DeepSeek-V3 to summarize and score the concepts. We use CLIP models from OpenCLIP~\cite{cherti2023reproducible} with ViT-L/14 as the default backbone. To train the linear head, we use Adam optimizer with the batch size of 64 and the learning rate of $5\times 10^{-5}$. All experiments were conducted on NVIDIA GeForce RTX 4090 GPUs. More details are provided in the Appendix.

\begin{figure*}[t]
\centering
\includegraphics[width=\textwidth]{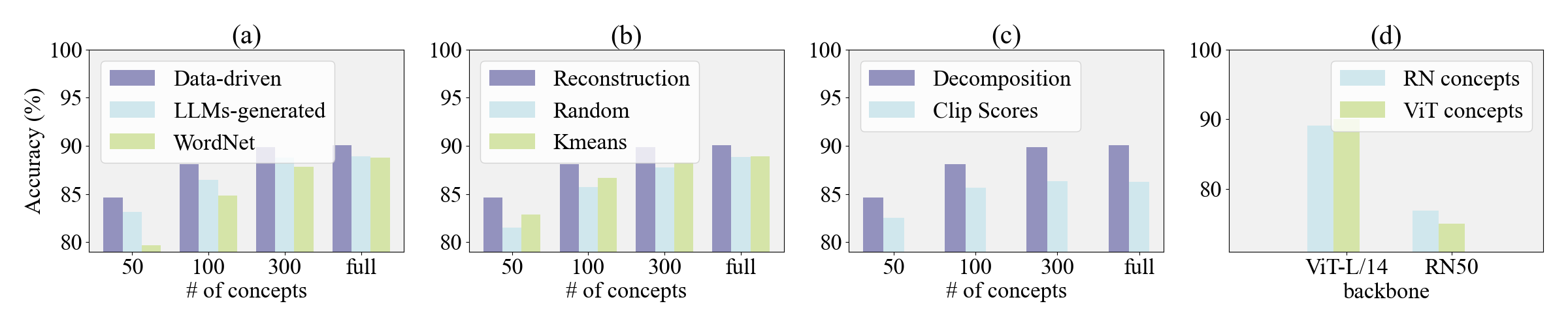}
\caption{Ablation studies. (a) Test accuracy with different creation schemes of the concepts and the size of the bottleneck. (b) Comparison between reconstruction-guided concept selection, random sampling, and K-Means. (c) Comparison between decomposition-based and CLIP similarity-based concept score association. (d) Effects of sources of the extracted concepts.}
\label{ablation}
\end{figure*}

\subsection{Main Results}
\noindent \textbf{Comparison with End-to-end Model.}
Table~\ref{tab:main1} presents the test accuracy of both PCBM-ReD and the linear probe. Our findings indicate that the performance difference between PCBM-ReD and the linear probe is minimal, with an average test accuracy that is just 0.41\% lower, despite improved interpretability. Notably, for several datasets, PCBM-ReD even outperforms the end-to-end model. In general, the performance of PCBM-ReD is affected by the concepts, which in turn depend on the quality of the description. For datasets like HAM, general MLLMs may struggle to accurately describe the skin tumors using specific terminology. Domain-specific MLLMs may further improve the performance.

\noindent \textbf{Comparison with other CBMs.} We also compare PCBM-ReD's performance with other CBMs in Table~\ref{tab:main1} and~\ref{tab:main2}. PCBM-ReD outperforms other language-guided CBMs such as LaBo and label-free CBM. It shows an average test accuracy that is 1.25\% higher than LaBo and 5.57\% higher than label-free CBM. It also achieves greater accuracy than CompDL and the original CBM, even though it does not depend on manually constructed concepts and annotations.

\subsection{Performance on Low-data Regime}
\noindent \textbf{Zero-shot Ability.}
Since the weighted sum of concept embeddings approximates the original image embeddings, it retains zero-shot capabilities similar to the original embeddings, something existing CBMs typically lack. In Table~\ref{tab:main1}, we present zero-shot accuracy using two strategies: the vanilla approach from CLIP and CuPL~\cite{pratt2023does}. For comparison, we include the performance of the original CLIP representation. The results show that our method achieves strong zero-shot performance, with average accuracy comparable to CLIP, while being interpretable.

\noindent \textbf{Few-shot Performance.}
We also evaluate the few-shot performance of PCBM-ReD across various datasets. We follow the few-shot evaluation protocol proposed by
CLIP with 1, 2, 4, 8, and 16 images randomly sampled
from the training set for each class. In Fig.~\ref{few_shot}, we present the average test accuracy, alongside the accuracy of LaBo for comparison. The full results can be found in Appendix. The results show PCBM-ReD consistently outperform LaBo. On average, PCBM-ReD surpasses LaBo by 5.01\%. This shows the potential application of our methods in few-shot learning.

\subsection{Interpretation Evaluation}
\noindent \textbf{Qualitative Analysis.}
To evaluate PCBM-ReD's interpretability, we present examples illustrating its reasoning process. In Fig.~\ref{explanation}, we visualize test set images alongside the top three concepts that most significantly contribute to the correct class's logit. The results show that PCBM-ReD effectively identifies key concepts in the images, which reliably predict corresponding labels. Furthermore, the concepts adapt to different data distributions. For general tasks like ImageNet classification, broad concepts (e.g., color, class) suffice to differentiate categories. In contrast, fine-grained classification requires more specific details (e.g., object parts). Additionally, the concepts are derived directly from visual features, which further enhances the interpretability.

\noindent \textbf{Human Evaluation.} We conduct a user study to evaluate the quality of the extracted concepts and generated explanations. Images from five datasets (Imagenet, Food-101, UCF-101, CUB-200-2011, and Flower-102) are randomly selected, and users are shown the image, prediction, and top concepts as explanations. Users rate the explanations on a scale of 1 to 5 based on: 1) whether the concepts are visually identifiable, 2) whether the concepts faithfully describe the image, and 3) whether the explanations have a clear causal link to class labels. Details of the study design are in the Appendix. For comparison, we evaluate explanations using concepts generated by prompting LLMs to describe the class based solely on prior knowledge without training image access~\cite{yang2023language}, and PCBM-ReD without LLM-based concept rating. Results from 39 volunteers (Fig.~\ref{user_study}) show that PCBM-ReD, which extracts concepts from image descriptions, produces mostly visually identifiable concepts. The concept rating step removes non-causal concepts, enabling the full PCBM-ReD to outperform all baselines.

\subsection{Analytical Studies}
We perform several ablation studies and report the average accuracy across CIFAR10, CIFAR100, and CUB.

\noindent \textbf{Bottleneck Size.}
We investigate how the model's performance changes with bottleneck size. As shown in Fig.~\ref{ablation} (a), test accuracy improves with more concepts in the bottleneck, nearly saturating at 300. Notably, even a small bottleneck of 50 achieves reasonable accuracy. Additionally, the required bottleneck size is independent of the number of label classes. Fewer concepts also enhance interpretability.

\noindent \textbf{Concept Creation Schemes.}
We compare our concept creation scheme with: (1) LLM-generated concepts~\cite{yang2023language} and (2) "Core" WordNet concepts~\cite{boyd2006adding}, using the same preprocessing and selection algorithms as PCBM-ReD. Results in Fig.~\ref{ablation} (a) show that our data-driven scheme outperforms both alternatives. Additionally, it avoids non-visual concepts, as confirmed by human evaluation, while also providing a performance boost.

\noindent \textbf{Concept Selection Methods.}
We evaluate the effectiveness of the reconstruction-guided concept selection algorithm by comparing it with two baselines: random sampling and k-means clustering. As shown in Fig.~\ref{ablation} (b), our method consistently outperforms both baselines, with a noticeable performance gap at smaller bottlenecks. By selecting concepts that minimize reconstruction error, the reconstruction-guided approach ensures higher final accuracy.

\noindent \textbf{Concept Score Association.}
We compare the sparsity decomposition-based concept score association and to baseline that uses CLIP similarity scores as concept scores. As shown in Fig.~\ref{ablation} (c), representation decomposition achieves significantly higher accuracy. By reconstructing visual representations with concepts, it ensures performance comparable to the original model, achieving higher accuracy.

\noindent \textbf{Sources of the Concepts.}
Finally, we examine the impact of concept sources by using concepts  extracted from different image encoders to construct CBMs. Results in Fig.~\ref{ablation} (d) show a performance decline when the backbones are mismatched, highlighting the importance of aligning concepts with the representational capabilities of the image encoder.
\section{Limitations and Conclusion}

In this paper, we propose PCBM-ReD, a novel CBM that generates concepts through post-hoc decomposition of visual representations. It achieves high accuracy, zero-shot/few-shot capabilities, and built-in interpretability. However, the approach has some limitations. First, the concept generation relies on MLLMs and LLMs, which can be suboptimal for domain-specific images that general MLLMs cannot describe precisely. Second, as a post-hoc method, the model's performance depends on the original image encoder. We will further improve the model by designing better prompts or leveraging more advanced MLLMs.
\section*{Acknowledgments}
This research work was supported in part by the National Natural Science Foundation of China (Project No. 62322318 and No. 62201485), in part by the Research Grants Council of Hong Kong Special Administrative Region, China (Project No. T45-401/22-N).

\bibliography{aaai2026}

\appendix
\section{Dataset Statistics}
Table~\ref{tab:datastats} depicts detailed statistics for all datasets.  We follow the same train/dev/test splits provided by LaBo~\cite{yang2023language}. For ImageNet, we only evaluate the dev set.

\section{Implementation Details}
\label{sec:implement}
\subsection{Linear Probe}
The test accuracy of the CLIP linear probe is referenced from~\cite{yang2023language}, which utilizes encoded images prior to their projection into the vision-text embedding space as inputs for the classifier. In our approach, we employ the approximated visual embedding—already situated in the vision-text embedding space—as the classifier's input. We implement logistic regression using sklearn's L-BFGS, allowing for a maximum of 1,000 iterations. To find the optimal values for the hyperparameter $C$, we perform a binary search on the validation set, starting with the range $[1e^6, 1e^4, 1e^2, 1, 1e^{-2}, 1e^{-4}, 1e^{-6}]$. After establishing the left and right bounds for $C$, we iteratively halve the interval over 8 steps to arrive at the final hyperparameter value.

\subsection{Training of Sparse Autoencoder}

We train sparse autoencoders following the setup of~\cite{bricken2023monosemanticity}. Specifically, we utilize the implementation from sparse autoencoder library. For each dataset, we train on the training data for
200 epochs, with expansion factor of 4. We perform hyperparameter sweeps
based on the $L_2$ reconstruction errors of the validation set over the learning rate with the range $[1e^{-5}, 5e^{-5}, 1e^{-4}, 5e^{-4}, 1e^{-3}]$ and $L_1$ sparsity coefficient with the range $[3e^{-5}, 1.5e^{-4}, 3e^{-4}, 1.5e^{-3}, 3e^{-3}]$.

\subsection{Sparse Decomposition}
 We utilize scikit-learn’s implementation of orthogonal matching pursuit for sparse decomposition. The orthogonal matching pursuit has a parameter, i.e., number of non-zero coefficients, which controls the sparsity of the decomposition. For the visual representation of each sample, we use binary search within the range $[650, \text{full}]$ to tune this parameter, based on the $L_2$ reconstruction error of the visual representation. Here ``full'' refers to the total number of concepts within the candidate pool. The specific hyper-parameters can be found in the released code.

\subsection{Prompts}
\label{sec:prompt}
\paragraph{Object Description.}
We ask the MLLMs to describe the objects with the following prompts. Specifically, we apply a two phase prompt. We first use the prompt-1, which ask the MLLMs to use chain-of-thought to determine the category of the object within the image. Then we use the prompt-2 to ask the MLLMs to summarize the reasoning process of the first stage, and give a concise description of the subject.

\begin{table}[tp]
\centering
\renewcommand\arraystretch{1}
\begin{center}
\resizebox{0.48\textwidth}{!}{
\begin{tabular}{ccccc}
\toprule
\multirow{2}{*}{\textbf{Name}} & \multirow{2}{*}{\textbf{n. of classes}} & \multicolumn{3} {c} {\textbf{n. of images} }\\
\cline{3-5}
& & Train & Dev & Test\\
\midrule
Food-101 & 101& 50,500& 20,200&30,300 \\
FGVC-Aircraft & 102&3,334&3,333&3,333\\
Flower-102 & 102 & 4,093 & 1,633&2,463\\
CUB-200-2011 & 200 & 3,994&2,000&5,794\\
UCF-101&101&7,639&1,898&3,783\\
DTD&47&2,820&1,128&1,692\\
HAM10000&7&8,010&1,000&1,005\\
RESISC45&45&3,150&3,150&25,200\\
CIFAR-10&10&45,000&5,000&10,000\\
CIFAR-100&100&45,000&5,000&10,000\\
ImageNet&1,000 & 1,281,167&50,000&-\\
\bottomrule
\end{tabular}}
\end{center}
    \caption{Detailed statistics of the 11 datasets. }
\label{tab:datastats}
\end{table}
\begin{itemize}
\item \textbf{Food-101}:
\begin{itemize}
\item prompt-1 \textsl{What is the food? Use chain of thought to reason and list all the visual features that lead to the conclusion.}
\item prompt-2 \textsl{Briefly summarize the appearance of the food in one paragraph. Do not mention what the food is.}
\end{itemize}
\item \textbf{FGVC-Aircraft}:
\begin{itemize}
\item prompt-1 \textsl{What is the model of the flight? Use chain of thought to reason and list all the visual features that lead to the conclusion.}
\item prompt-2 \textsl{Concisely summarize the appearance of the flight using one paragraph. Do not include the specific model name.}
\end{itemize}
\item \textbf{Flower-102}:
\begin{itemize}
\item prompt-1 \textsl{What is the species of the flower? Use chain of thought to reason and list all the visual features that lead to the conclusion.}
\item prompt-2 \textsl{Concisely summarize the appearance of the flower using one paragraph. Do not include the specific species name of the flower.}
\end{itemize}
\item \textbf{CUB-200-2011}:
\begin{itemize}
\item prompt-1 \textsl{What is the species of the bird? Use chain of thought to reason and list all the visual features that lead to the conclusion.}
\item prompt-2 \textsl{Concisely summarize the appearance of the bird using one paragraph. Do not include the specific species name of the bird.}
\end{itemize}
\item \textbf{UCF-101}:
\begin{itemize}
\item prompt-1 \textsl{What is the action the characters are performing? Use chain of thought to reason and list all the visual features that lead to the conclusion.}
\item prompt-2 \textsl{Briefly summarize visual pattern of the action using one paragraph. Do not mention what the action is.}
\end{itemize}
\item \textbf{DTD}:
\begin{itemize}
\item prompt-1 \textsl{What is the texture (e.g., blotchy, frilly)? Use chain of thought to reason and list all the visual features that lead to the conclusion.}
\item prompt-2 \textsl{Briefly summarize the feature of the texture in one paragraph, do not mention the name of the texture.}
\end{itemize}
\item \textbf{HAM10000}:
\begin{itemize}
\item prompt-1 \textsl{What is the type of the pigmented lesion in the dermatoscopic image (e.g., Actinic keratoses, basal cell carcinoma, benign keratosis-like lesions, dermatofibroma, melanoma, melanocytic nevi, and vascular lesions)? Use chain of thought to reason and list all the visual features that lead to the conclusion.}
\item prompt-2 \textsl{Briefly summarize the appearance of the pigmented lesion in one paragraph, do not include the specific type name.}
\end{itemize}
\item \textbf{RESISC45}:
\begin{itemize}
\item prompt-1 \textsl{What is the scene class in the satellite image? Use chain of thought to reason and list all the visual features that lead to the conclusion.}
\item prompt-2 \textsl{Briefly summarize the appearance of the scene in one paragraph. Do not mention what the scene is.}
\end{itemize}
\item \textbf{CIFAR-10}:
\begin{itemize}
\item prompt-1 \textsl{Describe the appearance of the object within the image. Only include the visual features that can help identify its category.  Use Chain-of-thought to reason.}
\item prompt-2 \textsl{Briefly summarize the appearance of the object in one paragraph. Do not mention what the object is.}
\end{itemize}
\item \textbf{CIFAR-100}:
\begin{itemize}
\item prompt-1 \textsl{Describe the appearance of the object within the image. Only include the visual features that can help identify its category.  Use Chain-of-thought to reason.}
\item prompt-2 \textsl{Briefly summarize the appearance of the object in one paragraph. Do not mention what the object is.}
\end{itemize}
\item \textbf{ImageNet}:
\begin{itemize}
\item prompt-1 \textsl{Describe the appearance of the object within the image. Only include the visual features that can help identify its category.  Use Chain-of-thought to reason.}
\item prompt-2 \textsl{Briefly summarize the appearance of the object in one paragraph. Do not mention what the object is.}
\end{itemize}
\end{itemize}

\begin{figure*}[t]
\centering
\includegraphics[width=\textwidth]{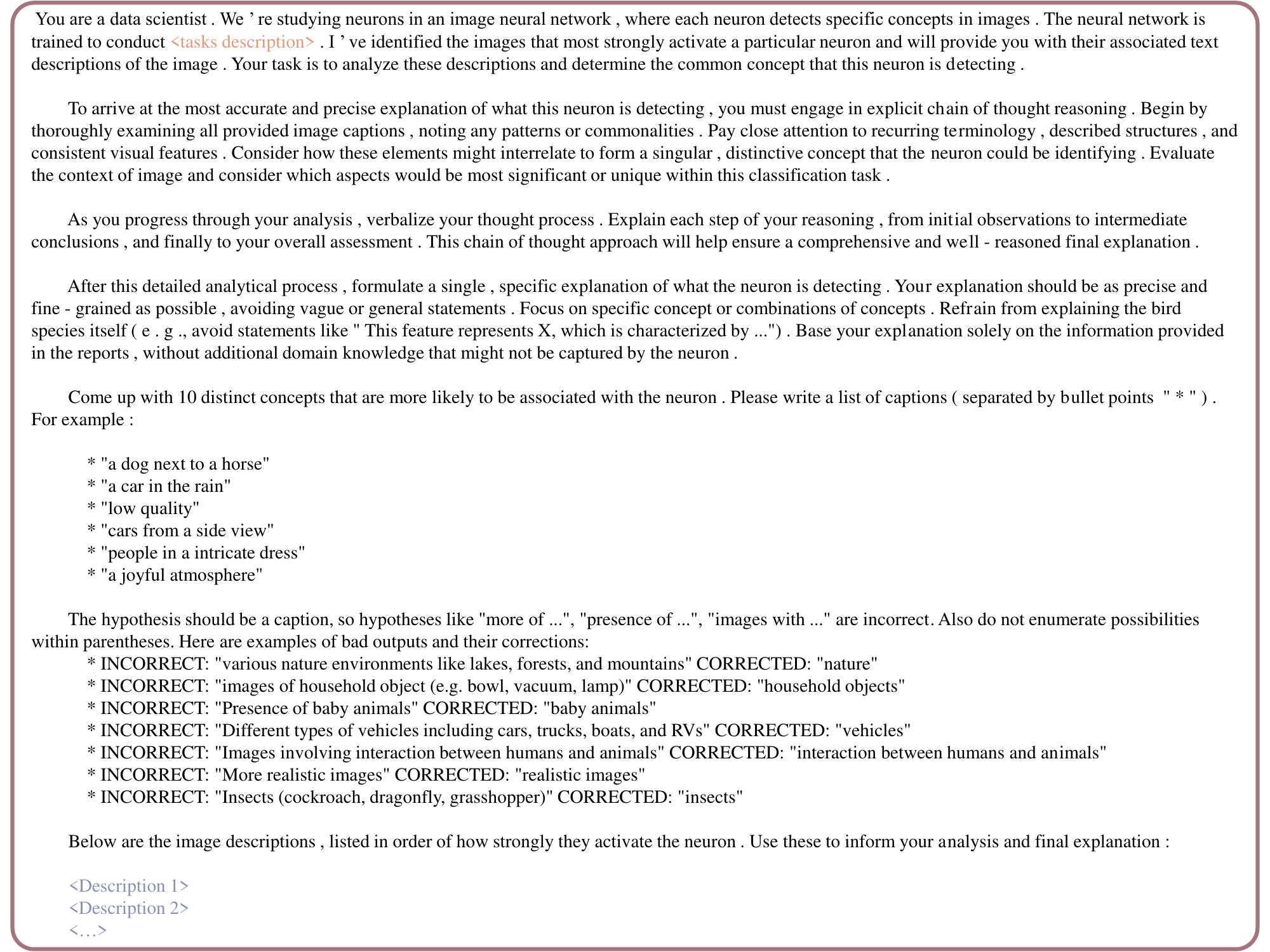}
\caption{Prompt used for summarizing the descriptions and generating candidate concepts.}
\label{prompt1}
\end{figure*}

\paragraph{Concept Summarization.}
Once we have descriptions for each image, we gather the descriptions for the top 20 activations in each potential concept. We then utilize the prompt shown in Fig.~\ref{prompt1} to request LLMs to summarize these descriptions and suggest candidate concepts. The tasks description is tailored according to the specific dataset (e.g., for CUB-200-2011, the task description is ``bird species classification''). Finally, we follow~\cite{yang2023language} to replace the specific name of the category by its higher level class name (e.g., specific bird species name is replaced with ``bird'').

\begin{figure*}[t]
\centering
\includegraphics[width=\textwidth]{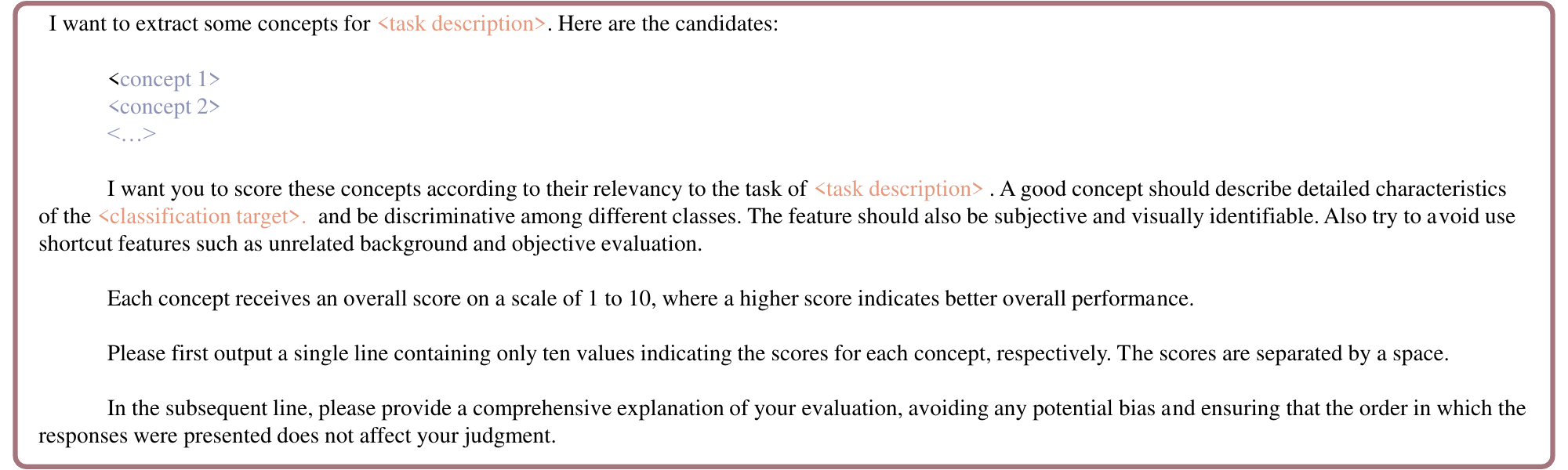}
\caption{Prompt used for rating the candidate concepts.}
\label{prompt2}
\end{figure*}

\paragraph{Concept Rating.} Once we obtain a pool of candidate concepts, we ask LLMs to rate each concepts according to their relevancy to the specific tasks. The prompt we use is presented in Fig.~\ref{prompt2}. Each candidate is given a score between 1 to 10 according to its quality. We only keep concepts with score greater than 6 for the subsequent processing.

\subsection{Human Evaluation}
\label{sec:user}
Here we describe the details how we design the user study to evaluate the interpretability of the proposed PCBM-ReD and the baselines.
\paragraph{Method.} To perform human evaluation, we randomly sample several images from the test set of ImageNet~\cite{deng2009imagenet}, Food-101~\cite{bossard2014food}, UCF-101~\cite{soomro2012ucf101}, CUB-200-2011~\cite{wah2011caltech}, and Flower-102~\cite{nilsback2008automated}. The samples selected all have correct prediction by the corresponding methods. We select these five datasets because they are common objects that can be evaluated with common knowledge. For each image, we use either PCBM-ReD, PCBM-ReD without the LLMs-based concept scoring, or LaBo~\cite{yang2023language} to generate explanations. We generate explanations for multiple images, i.e., different users will see different cases. 

\begin{figure*}[t]
\centering
\includegraphics[width=\textwidth]{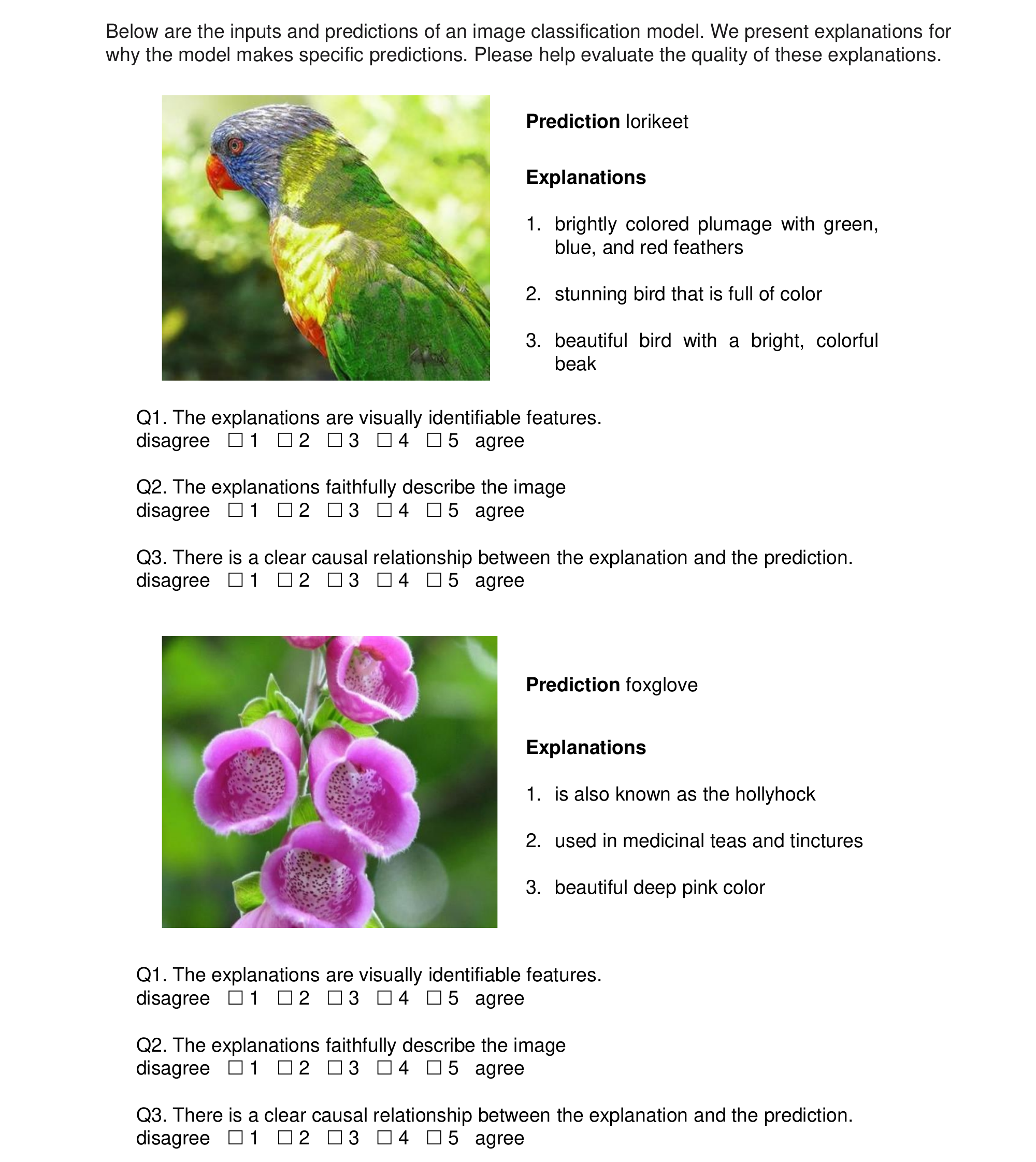}
\caption{Examples of questions in the user study. For each volunteer, we show the image, the prediction, and the corresponding explanations. We ask the user to select whether they agree with certain statements.}
\label{fig:survey}
\end{figure*}

\paragraph{Question Structure.} For each question, we show the volunteer three statement, and ask them to select whether they agree with the statement. The statement includes: 1) The explanations are visually identifiable features, 2) The explanations faithfully describe the image, 3) There is a clear causal relationship between the explanation and the prediction. We ask the volunteers to rate on a scale of 1 and 5, with 1 to be very disagree with the statement and 5 to be very agree with the statement.
\paragraph{Survey Structure.} For each volunteers, we randomly sample 2 methods out of 3 methods, and show the images and the corresponding explanations sampled from 5 datasets. The order of the images are also shuffled. Each volunteer will see 10 images and answer 30 questions in total. Fig.~\ref{fig:survey} shows a sample survey. We published the survey
internally, and received 39 responses.

\paragraph{Ethic Statement} In our user study, we did not collect any user information. The IRB review is waived because the study does not involve human subjects as defined in 45 CFR 46.102(f).

\section{Derivation of Alg. 1}
\label{sec:algo}
In this section, we provide more explanation on the Alg.~\ref{alg:alg}. We start with the optimization objective:
\begin{equation}
\min_\mathcal{C} \sum_{i=1}^N \min_{\mathbf{\beta}_i(\mathcal{C})} \Vert \mathbf{I}_i - \mathbf{R}(\mathcal{C})^T\mathbf{\beta}_i(\mathcal{C}) \Vert_F^2.
\end{equation}
We notice the inner minimization problem has analytical solution:
\begin{equation}
\beta^*_i(\mathcal{C}) = (\mathbf{R}(\mathcal{C})\mathbf{R}(\mathcal{C})^T)^{-1}\mathbf{R}(\mathcal{C})\mathbf{I}_i.
\end{equation}
The greedy algorithm works by incrementally adding new concept into $\mathcal{C}$. Assume we already have $n$ element in $\mathcal{C}_n$. At the $n+1$ step, we will find the optimal concept $c^*$ that minimize the objective:
\begin{equation}
\begin{split}
\min_c \sum_{i=1}^N &\Vert \mathbf{I}_i - \mathbf{R}(\mathcal{C}_n\cup \{c\})^T
\\&(\mathbf{R}(\mathcal{C}_n\cup \{c\})\mathbf{R}(\mathcal{C}_n\cup \{c\})^T)^{-1}\mathbf{R}(\mathcal{C}_n\cup \{c\})\mathbf{I}_i \Vert_F^2.
\end{split}
\end{equation}

A vanilla way to calculate $c^*$ is to enumerate all $c\in\mathcal{C}_0$ and compute the objective value, and select the $c^*$ that minimize the objectives. However, this can be computational expensive especially when the size of  $\mathcal{C}_0$ is very large. However, we notice that 
\begin{align*}
&\mathbf{R}(\mathcal{C}_n\cup \{c\})^T(\mathbf{R}(\mathcal{C}_n\cup \{c\})\mathbf{R}(\mathcal{C}_n\cup \{c\})^T)^{-1}\mathbf{R}(\mathcal{C}_n\cup \{c\})\\
=&[\mathbf{R}(\mathcal{C}_n)^T,\mathcal{T}(c)]\begin{bmatrix}
\mathbf{R}(\mathcal{C}_n)\mathbf{R}(\mathcal{C}_n)^T & \mathbf{R}(\mathcal{C}_n)\mathcal{T}(c)\\
\mathcal{T}(c)^T\mathbf{R}(\mathcal{C}_n)^T &\mathcal{T}(c)^T\mathcal{T}(c)
\end{bmatrix}^{-1}\begin{bmatrix}
\mathbf{R}(\mathcal{C}_n)\\
\mathcal{T}(c)^T
\end{bmatrix}
\end{align*}
By defining $\mathbf{A} = (\mathbf{R}(\mathcal{C}_n)^T\mathbf{R}(\mathcal{C}_n))^{-1}$, $\mathbf{P} = \mathbf{R}(\mathcal{C}_n)(\mathbf{R}(\mathcal{C}_n)^T\mathbf{R}(\mathcal{C}_n))^{-1}\mathbf{R}(\mathcal{C}_n)^T$ and $z=\mathcal{T}(c)^T\mathcal{T}(c)-\mathcal{T}(c)^TP\mathcal{T}(c)$, we have:
\begin{align*}
&\begin{bmatrix}
\mathbf{R}(\mathcal{C}_n)\mathbf{R}(\mathcal{C}_n)^T & \mathbf{R}(\mathcal{C}_n)\mathcal{T}(c)\\
\mathcal{T}(c)^T\mathbf{R}(\mathcal{C}_n)^T &\mathcal{T}(c)^T\mathcal{T}(c)
\end{bmatrix}^{-1}\\
=&\begin{bmatrix}
\mathbf{A} + \mathbf{A}\mathbf{R}(\mathcal{C}_n)\mathcal{T}(c)\mathcal{T}(c)^T\mathbf{R}(\mathcal{C}_n)^T\mathbf{A}/z & -\mathbf{A}\mathbf{R}(\mathcal{C}_n)\mathcal{T}(c)/z\\
-\mathcal{T}(c)^T\mathbf{R}(\mathcal{C}_n)^T\mathbf{A}/z & 1/z
\end{bmatrix}.
\end{align*}
Therefore, we find that we do not need to recalculate the matrix inverse for every $c\in \mathcal{C}_0$, which is the most time-consuming operation. Instead, we only need to calculate $\mathbf{A} = (\mathbf{R}(\mathcal{C}_n)^T\mathbf{R}(\mathcal{C}_n))^{-1}$ once, and the result can be reused for all $c$. By further simplifying the formula, we have
\begin{align*}
    &\mathbf{R}(\mathcal{C}_n\cup \{c\})^T(\mathbf{R}(\mathcal{C}_n\cup \{c\})\mathbf{R}(\mathcal{C}_n\cup \{c\})^T)^{-1}\mathbf{R}(\mathcal{C}_n\cup \{c\})\\
    =&\mathbf{P}\mathbf{Q}\mathbf{P}-\mathbf{Q}\mathbf{P}-\mathbf{P}\mathbf{Q}+\mathbf{P}+\mathbf{Q},
\end{align*}
where $\mathbf{Q}=\mathcal{T}(c)\mathcal{T}(c)^T/z$. This gives the exact algorithm as Alg. 1. For each $c$, we reuse the common $\mathbf{P}$ and recalculate $\mathbf{Q}$. The calculation of $\mathbf{Q}$ only entails matrix multiplication and scaler division, being more efficient than the original formula which entails matrix inversion. Note that the simplification holds if and only if $z\neq0$. The condition $z=0$ means there is linear dependency between $c$ and existing elements within $\mathcal{C}_n$. To guarantee the selected concepts are independent, we should remove such $c$. 

\section{Additional Results}
\label{sec:result}
\subsection{Full Few-shot Performance Result}
\begin{figure*}[t]
\centering
\includegraphics[width=\textwidth]{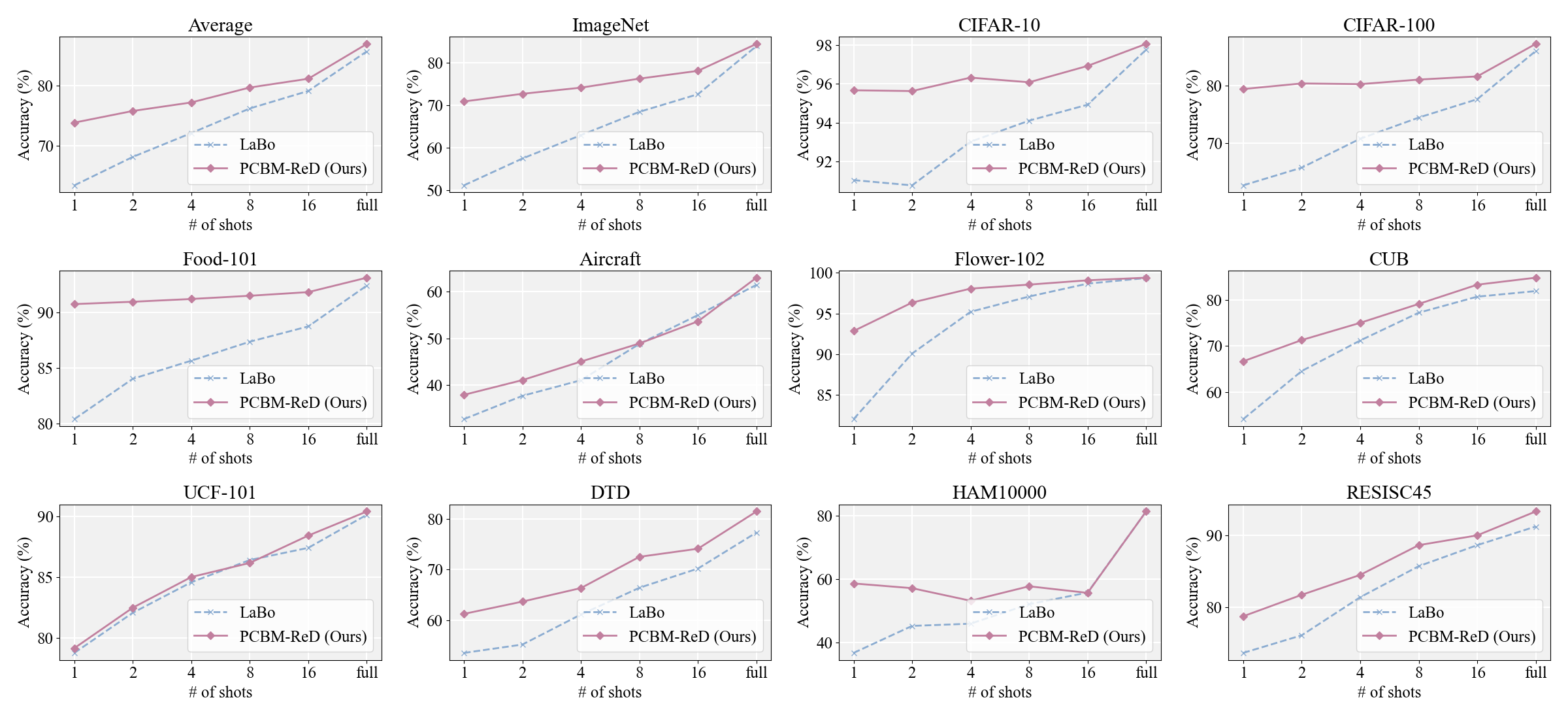}
\caption{Few-shot test accuracy (\%) comparison between PCBM-ReD and LaBo on 11 datasets. The x-axis represets the number of labeled images for each class.}
\label{fig:few_shot}
\end{figure*}

We present the full few-shot performance result in Fig.~\ref{fig:few_shot}. We show the test accuracy of 1, 2, 4, 8, and 16-shot on 11 datasets. The shot refers to the labeled image per classes during training phase. Since the concept extraction and selection stages are conducted in an unsupervised manner, we still utilize all the training data for concept extraction and selection (i.e., the concepts used for building CBMs are the same as that for fully-supervised setting). The results demonstrate that PCBM-ReD consistently outperforms baselines across different tasks and settings.

\subsection{Complete Results of Human Evaluation} 
\begin{figure*}[t]
\centering
\includegraphics[width=\textwidth]{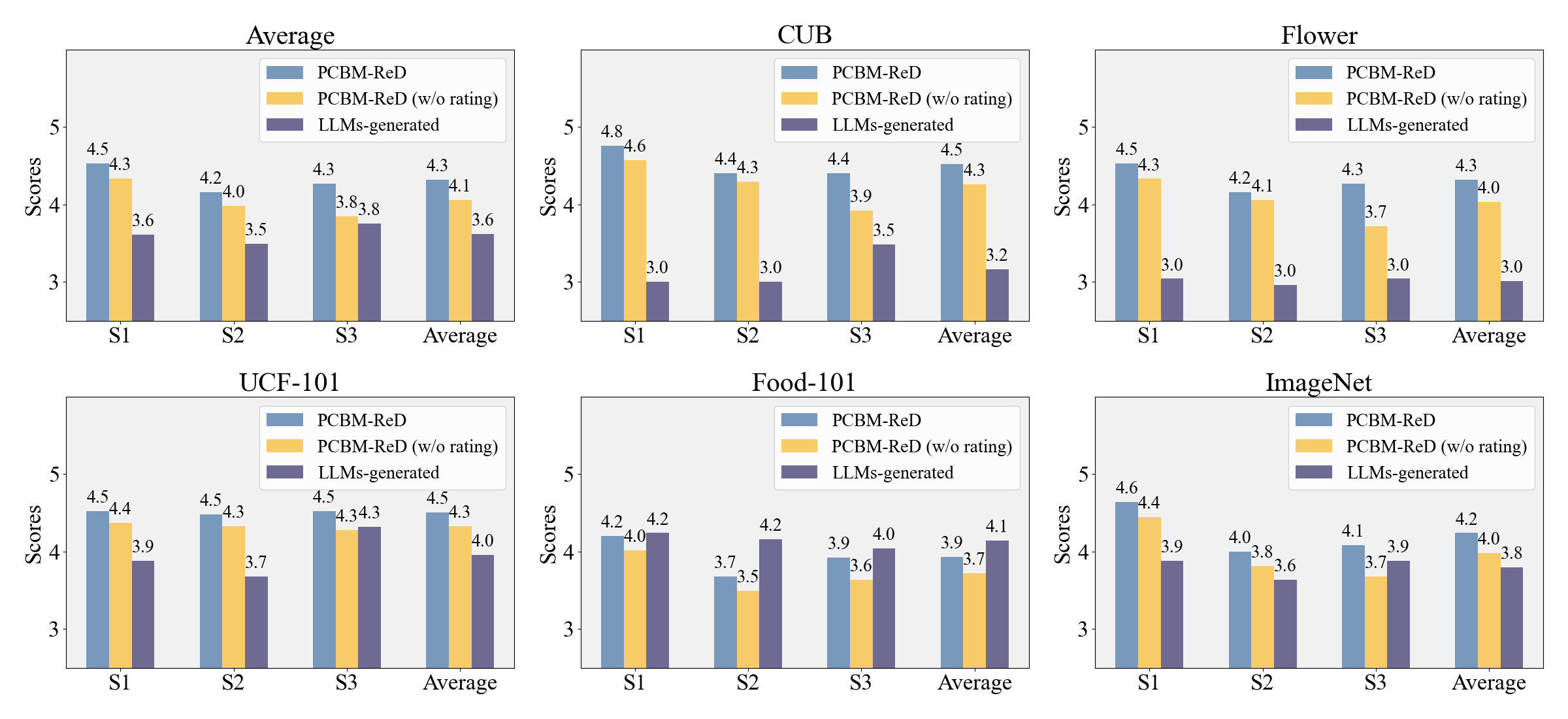}
\caption{Full results of user studies. We show the average scores for 5 individual datasets.}
\label{user_a}
\end{figure*}

In Fig.~\ref{user_a}, we present the complete human evaluation results. We observe that, for most datasets, PCBM-ReD significantly outperforms LaBo. However, there are still instances of failure. For example, in the case of the food dataset, PCBM-ReD achieved lower scores than LaBo. This is likely because MLLMs struggle to describe the visual patterns of food without explicitly naming it, especially when the ingredients have been processed and is hard to distinguish. In contrast, LLMs tend to describe food based on its ingredients and cooking methods, resulting in more accurate concepts.

\subsection{Additional Qualitative Result}
\begin{figure*}[t]
\centering
\includegraphics[width=\textwidth]{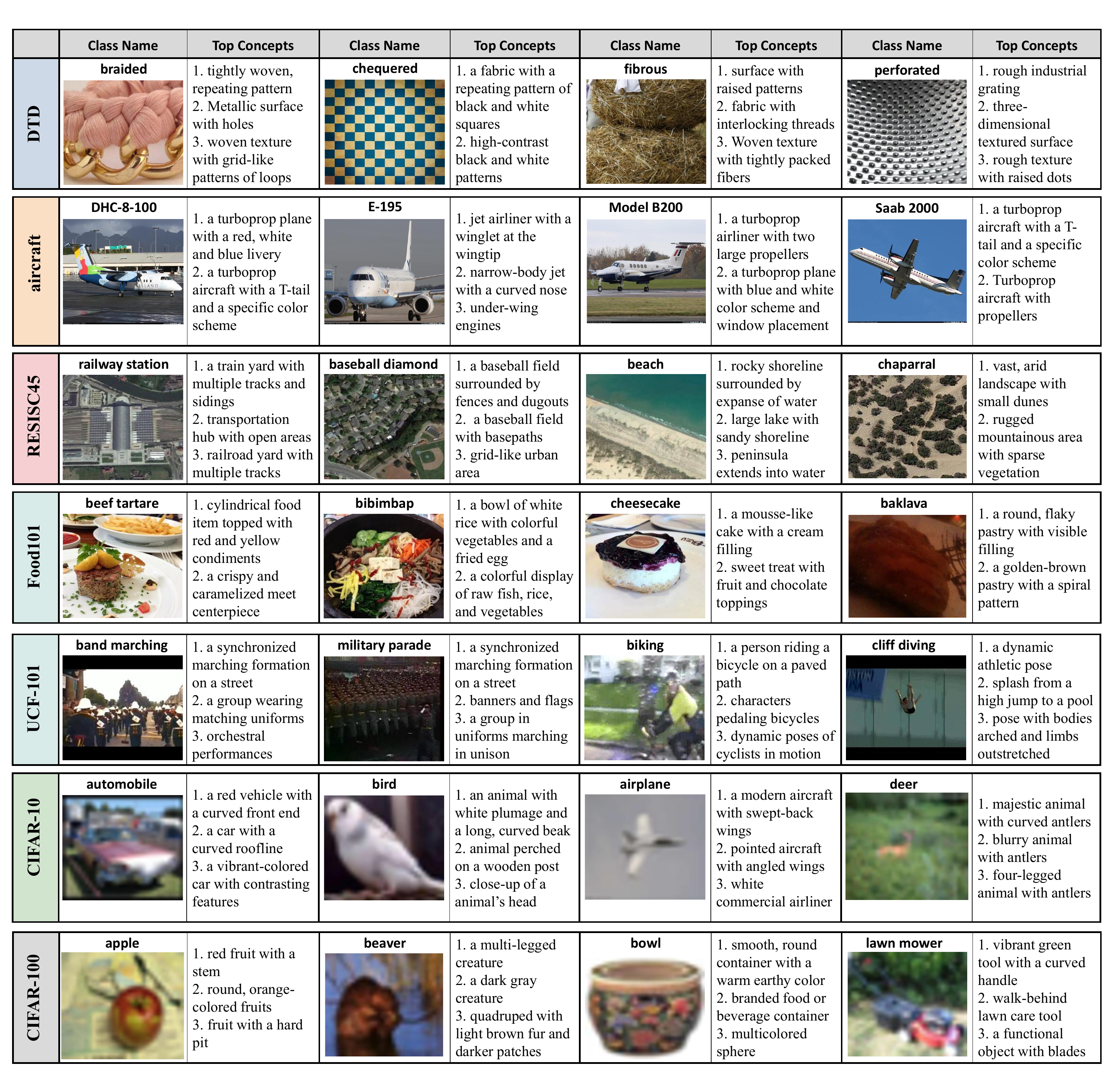}
\caption{Several example explanations generated by PCBM-ReD. The examples are sampled from the test set of 7 datasets uncovered by the main paper, which have correct predictions. We also show their corresponding top concepts that contribute the most to the logits.}
\label{exp_a}
\end{figure*}

In this section, we present additional qualitative examples for the remaining six datasets in Fig.~\ref{exp_a}. Our method offers coherent explanations for its predictions.

\subsection{Statistical Significance}
To verify the improvements of PCBM-ReD over baselines, we perform statistical test.  We perform Wilcoxon signed-rank (WSR) tests on the accuracy over 11 datasets, yielding p-values of 0.005 (vs. Labo), 0.001 (vs. Res-CBM), and 0.002 (vs. V2C-CBM), showing statistically significant improvements over baselines. For the results of user study shown in Fig.~\ref{user_study}, we also perform WSR test and the result shows $p<0.05$ for all pairs.

\subsection{Extension to non-CLIP Model}

The PCBM-ReD framework can be extended to non-CLIP models with simple modifications, as its core components: concept extraction, concept selection, and representation decomposition, are not tied to CLIP. For non-CLIP models, techniques like concept activation vectors (CAVs)~\cite{kim2018interpretability} can be used to construct concept representations, bypassing the need for a text encoder. Given a candidate concept and captions for several images, we can prompt the LLM to determine whether each image contains the concept. This results in multiples positive and negative samples. These samples are then used to derive CAVs as the concept representation. \citet{gong2025concepts} gives details of non-CLIP extesion of PCBM-ReD.

\end{document}